\documentclass[10pt,twocolumn,letterpaper]{article}

\usepackage{iccv}
\usepackage{times}
\usepackage{epsfig}
\usepackage{graphicx}
\usepackage{amsmath}
\usepackage{amssymb}

\usepackage{relsize}
\usepackage{url}
\usepackage{tabularx, booktabs}
\usepackage{caption}
\usepackage{pifont}
\usepackage[table]{xcolor}
\definecolor{lightgray}{gray}{0.9}
\definecolor{lightblue}{rgb}{0.93,0.95,1.0}
\definecolor{darkgreen}{rgb}{0.0,0.6,0.0}
\definecolor{mypink1}{rgb}{0.858, 0.188, 0.478}

\newcommand\roeih[1]{\textcolor{blue}{[RH: #1]}}

\newcommand\huijuan[1]{\textcolor{cyan}{[HX: #1]}}

\newcommand{\ignore}[1]{}

\newcolumntype{L}[1]{>{\raggedright\let\newline\\\arraybackslash\hspace{0pt}}m{#1}}
\newcolumntype{C}[1]{>{\centering\let\newline\\\arraybackslash\hspace{0pt}}m{#1}}
\newcolumntype{R}[1]{>{\raggedleft\let\newline\\\arraybackslash\hspace{0pt}}m{#1}}
\renewcommand{\eqref}[1]{Eq.~\ref{#1}}

\newcommand{\figref}[1]{Fig.~\ref{#1}}
\newcommand{\tabref}[1]{Tab.~\ref{#1}}
\newcommand{\secref}[1]{Sec.~\ref{#1}}

\newcommand{\reals}{\mathbb{R}}
\newcommand{\vect}[1]{\boldsymbol{\mathbf{#1}}}

\def\beq{\begin{equation}}
\def\eeq{\end{equation}}

\def\beqary{\begin{eqnarray}}
\def\eeqary{\end{eqnarray}}

\def\beqarz{\begin{eqnarray*}}
\def\eeqarz{\end{eqnarray*}}

\usepackage[pagebackref=true,breaklinks=true,letterpaper=true,colorlinks,bookmarks=false]{hyperref}

\iccvfinalcopy 

\def\iccvPaperID{21} 

\ificcvfinal\fi
\usepackage{boxedminipage}
\usepackage{bbold}

\renewcommand{\xi}{{\xx}^{(m)}}

\newcommand{\cV}{\mathcal{V}}

\newcommand{\needcite}[1]{}

\newcommand{\be}{\begin{equation}}
\newcommand{\ee}{\end{equation}}
\newcommand{\benn}{\begin{equation*}}
\newcommand{\eenn}{\end{equation*}}
\newcommand{\bea}{\begin{eqnarray*}}
\newcommand{\eea}{\end{eqnarray*}}
\newcommand{\bean}{\begin{eqnarray}}
\newcommand{\eean}{\end{eqnarray}}

\newcommand{\xx}{\boldsymbol{x}}

\newcommand{\zz}{\boldsymbol{z}}

\newcommand{\vv}{\boldsymbol{v}}

\newcommand{\ve}{\zz}

\newcommand{\comment}[1]{}

\newcommand{\polyring}[1]{\reals\left[x_1,\ldots,x_n\right]}

\definecolor{atomictangerine}{rgb}{0.8, 0.2, 0.1}
\definecolor{turq}{rgb}{0.0, 0.5, 0.5}
\definecolor{darkturq}{rgb}{0.0, 0.4, 0.4}
\definecolor{bright}{rgb}{0.8, 0.1, 0}
\definecolor{darkgray}{gray}{0.3}
\definecolor{mahogany}{rgb}{0.6, 0.05, 0.05}
\definecolor{pink}{rgb}{1,0.05,0.6}
\definecolor{myblue}{rgb}{0.3,0.05,0.9}

\newcommand\ag[1]{\textcolor{blue}{(\textbf{AG:} #1 )}}

\renewcommand{\eqref}[1]{Eq.~\ref{#1}}

\begin{document}

\title{Spatio-Temporal Action Graph Networks}

\author{
Roei Herzig$^{1^{\star, \dagger}}$, \,\,
Elad Levi$^{2^\star}$, \,\,
Huijuan Xu$^{3^\star}$, \,\,
Hang Gao$^3$, \,\,
Eli Brosh$^2$, \,\, \\
Xiaolong Wang$^3$, \,\,
Amir Globerson$^1$, \,\,
Trevor Darrell$^{2,3}$ \vspace{3pt}\\
$^1$Tel Aviv Univeristy, $^2$Nexar, $^3$UC Berkeley\\
}

\maketitle

\renewcommand*{\thefootnote}{$\star$}
\setcounter{footnote}{1}
\footnotetext{Equal Contribution.}

\renewcommand*{\thefootnote}{$\dagger$}
\setcounter{footnote}{2}
\footnotetext{Work done during an internship at Nexar.}

\renewcommand*{\thefootnote}{\arabic{footnote}}
\setcounter{footnote}{0}
\thispagestyle{empty}

\vspace{-10pt}
\begin{abstract}
\vspace{-10pt}
\label{sec:abstract}
Events defined by the interaction of objects in a scene are often of critical importance; yet important events may have insufficient labeled examples to train a conventional deep model to generalize to future object appearance.   
Activity recognition models that 
represent object interactions explicitly have the potential to learn in a more efficient manner than those that represent scenes with global descriptors. 
We propose a novel inter-object graph representation for activity recognition based on a disentangled graph embedding with direct observation of edge appearance. 
In contrast to prior efforts, our approach uses explicit appearance for high order relations derived from object-object interaction, formed over regions that are the union of the spatial extent of the constituent objects.
We employ a novel factored embedding of the graph structure, disentangling a representation hierarchy formed over spatial dimensions from that found over temporal variation. 
We demonstrate the effectiveness of our model on the Charades activity recognition benchmark, as well as a new dataset of driving activities focusing on multi-object interactions with near-collision events. Our model offers significantly improved performance compared to baseline approaches without object-graph representations, or with previous graph-based models.
\vspace{-15pt}
\end{abstract}


\section{Introduction}
\label{sec:introduction}

\begin{figure}[t!]
	\begin{center}
        \includegraphics[width=0.90\linewidth]{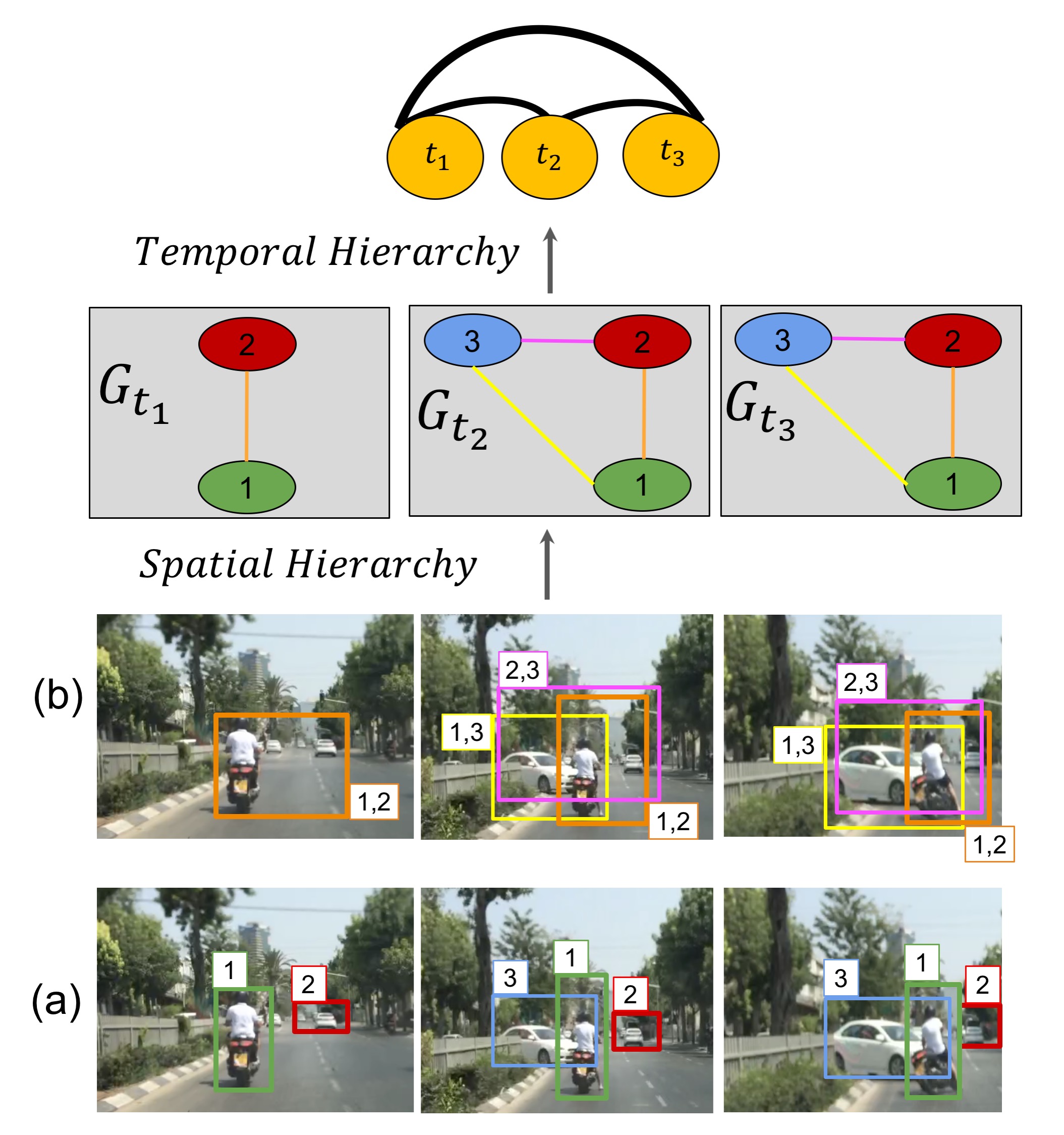}
    \vspace{-10pt}
	\caption{\small{Our  Spatio-Temporal Action Graph (STAG) approach models inter-object relations per frame. Each frame representation contains objects (a) and their relations (b). Our model contains two hierarchies: a spatial hierarchy to model all possible object interactions in a frame and a temporal hierarchy to aggregate the temporal context for the whole video.}}
	\vspace{-25pt}
	\label{fig:tesear}
	\end{center}
\end{figure}



Recognition of events in natural scenes poses a challenge for deep learning approaches to activity recognition, since an insufficient number of training examples are typically available to learn to generalize to all required observation conditions and variations in appearance. For example, in driving scenarios critical events are often a function of the spatial relationship of prominent objects, yet available event training data may not exhibit variation across a sufficiently wide range of appearances.
%
%
E.g., if a conventional deep model has only seen red pickup trucks rear-end blue sedans, and green trucks always drive safely in the training set, it may perform poorly in a test condition when observing a green pickup truck that is actually about to hit a red sedan.  
It is thus important to develop activity recognition models that can generalize effectively across object appearance and inter-object interactions.  

Early deep learning approaches to activity recognition were limited to scene-level image representations, directly applying convolutional filters to full video frames, and thus not modeling objects or their interactions explicitly \cite{donahue2015long,tran2015learning,simonyan2014two}.
Although networks are growing deeper and wider, the method for extracting features from network backbones is often still a basic pooling operation, with or without a pixel-wise attention step. These conventional deep learning approaches are unable to directly attend to objects or their spatial relationships explicitly.

A natural approach to the above problem is to build models that can capture relations between objects across time. This object-centric approach can decouple the object detection problem (for which more data is typically available. e.g., images of cars) and the problem of activity recognition. 
Many classic approaches to activity recognition explored object-based representations  \cite{gkioxari2015finding,weinzaepfel2016towards,mettes2017spatial,packer2012combined, brendel2011learning,gupta2009observing}; yet with conventional learning methods such approaches did not show significant improvements in real-world evaluation settings.
Several deep models have been recently introduced that directly represent objects in video in activity recognition tasks. For example \cite{baradel2018object} uses a relation network followed by an RNN, and \cite{wang2018videos} uses a spatio-temporal graph constructed whose nodes are detected objects. These models showed that a deep model with a dense graph defined over scene elements can lead to increased performance, but were limited in that only unary object appearance was considered, with a fully-connected spatio-temporal graph without taking object relations into account.


In this paper we propose a novel Spatio-Temporal Action Graph (STAG), which offers improved activity recognition. Our model design is motivated by the following two points. First, we observe that relations between objects are captured in the bounding box containing both objects more effectively than in the object boxes individually. Our graph utilizes explicit appearance terms for edges in the graph, forming a type of ``visual phrase'' term for each edge~\cite{visualphrases}: edge weights in our graph are formed using a descriptor pooled over the spatial extent of the union of boxes of each object pairs. Our experiments prove that modeling the visual appearance between objects outperforms other techniques (e.g. similarity and concatenation) for object-object interactions.


Second, the object interactions in one video are more concentrated in certain times which requires more structured spatial-temporal hierarchical feature representation. We propose a spatio-temporal disentangled feature embedding in our graph, factoring spatial and temporal connections into two hierarchies which first refine the edges considering all possible relations in a frame over space, and then over time. In our spatial hierarchy, the relations are refined by considering all possible relations within a frame and then aggregated to form per-frame descriptor. Next, we use the temporal hierarchy to aggregate the temporal context for the whole video and use it as input to the video classifier. We argue that this architecture is ideally structured for capturing relations that underlie typical actions in video. Indeed our empirical results show that it outperforms other spatio-temporal approaches without explicit hierarchy, including LSTM based ones.

Another key contribution of our work is a new dataset\footnote{The publicly available dataset can be found at: \href{https://github.com/roeiherz/STAG-Nets}{https://github.com/roeiherz/STAG-Nets}.} for collision activity detection in driving scenario. Activity recognition is of key importance in the domain of autonomous driving, in particular detecting collisions or near-collisions is of utmost importance. Most of the research on this topic is in simulation mode, and we introduce the first attempt to studying it in real world data. Additionally, this is the first dataset containing object-object interactions, while the current activity recognition datasets \cite{goyal2017something,sigurdsson2018charades} mostly contain limited are human-object interactions that have small number of objects per scene. Thus, they cannot contain rich relation information. Here we provide a new dataset which will allow researchers to study recognition of such rare and complex events. Our Collision dataset was collected from real-world dashcam data consisting of 803 videos containing collisions or near-collisions from more than ten million rides.\footnote{There are only relatively few collision videos, since naturally, such events are rare.} In \figref{fig:tesear}, we demonstrate our approach on a driving scene for collision event detection.

We evaluate our STAG model on both the Collision dataset as well the Charades~\cite{sigurdsson2016hollywood} activity recognition benchmark, demonstrating improvement over previous baselines. Our results confirm that the use of explicit object representations in spatial-temporal hierarchy can offer better generalization performance for deep activity recognition in realistic conditions with limited training data.

\vspace{-10pt}
\section{Related Work}
\vspace{-5pt}
\label{sec:related_work}

\begin{figure*}[t!]
    \begin{center}
    \includegraphics[width=\linewidth]{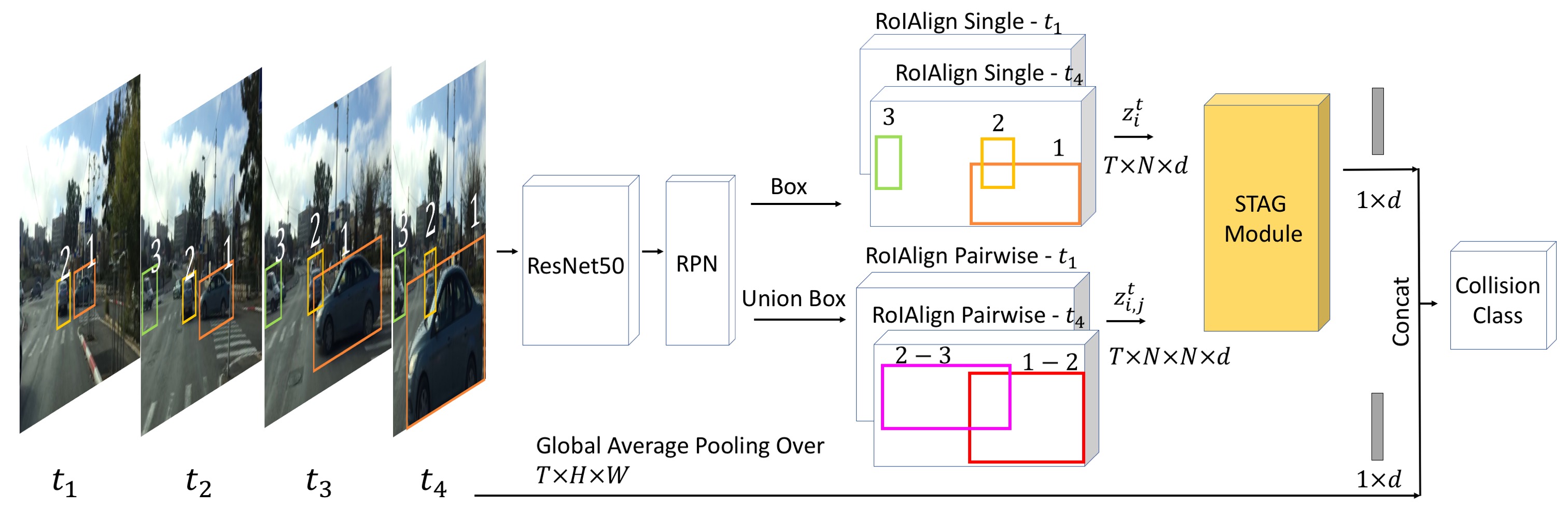}
    \vspace{-15pt}
    \caption{\textbf{STAG Network Architecture}. 
    Our STAG network architecture is comprised of: (1) A backbone of ResNet50 and RPN produces a set of bounding boxes proposals for each frame. (2) An \textit{RoIAlign Single} layer extracts $\ve_{i}^{t}$ features from the backbone for the boxes. In parallel, every pair of box proposals is used for computing a union box, and pairwise features $\ve_{i,j}^{t}$ are extracted similar to the $\ve_{i}^{t}$ features. (3) The $\ve_{i}^{t}$ and $\ve_{i,j}^{t}$ are used as inputs to an Spatio-Temporal Action Graph  Module (see \figref{fig:actions_graph}) which outputs a $d$ dimensional feature representing the entire sequence.}
    \vspace{-25pt}
    \label{fig:system}
    \end{center}
\end{figure*}

\textbf{Video Activity Recognition.} 
Early deep learning activity recognition systems were essentially ``bag of words models'', where per-frame features are pooled over an entire video sequence~\cite{karpathy2014large}. 
Later work used recurrent models (e.g., LSTM), to temporally aggregate frame features~\cite{donahue2015long,yue2015beyond}. 
Another line of activity classifiers use 3D spatio-temporal filters to create hierarchical representations of the whole input video sequence~\cite{ji20133d,taylor2010convolutional,tran2015learning,varol2018long}. While spatio-temporal filters can benefit from large video datasets, they cannot be pretrained on images. Two-stream networks were proposed to leverage image pretraining in RGB as well as capturing fine low-level motion in another optical flow stream~\cite{simonyan2014two,feichtenhofer2016convolutional}. I3D \cite{Carreira2017QuoVA} was designed to inflate 2D kernels to 3D to learn spatio-temporal feature extractors from video while leveraging image pre-trained weights. In all these models, whole frame video features are extracted without using object and inter-object details as our model does.

Object interactions have been utilized for tackling various activity recognition tasks~\cite{li2007and,yao2012recognizing,sun2018actor,girdhar2018video,moore1999exploiting,gupta2009observing}, e.g. by using spatio-temporal tubes \cite{gkioxari2015finding,weinzaepfel2016towards}, spatially-aware embeddings \cite{mettes2017spatial}, and spatio-temporal graphical models \cite{packer2012combined, brendel2011learning,gupta2009observing, girdhar2019video}. Probabilistic models have also been used in this context~\cite{ryoo2007hierarchical,fathi2011understanding,wu2007scalable}. But with conventional learning methods the addition of explicit object detection models often did not show significant improvements in real-world evaluations. Recently, object interactions in adjacent frames were modeled in \cite{ma2017attend,baradel2018object} followed by RNNs for capturing temporal structure. Also, Wang \cite{Wang_videogcnECCV2018} proposes to represent videos as space-time region graphs and perform reasoning on this graph representation via Graph Convolutional Networks. In \cite{Wang_videogcnECCV2018} objects in one video are allowed to interact with each other without constraints while we enforce more structured spatial-temporal feature hierarchy for better video feature encoding.


\textbf{Graph Neural Networks and Self-attention.}
Recently graph neural networks have been successfully applied in many computer vision applications: visual relational reasoning~\cite{battaglia2018relational,zambaldi2018relational,baradel2018object, latent_sgs_rr, referential_relationships}, image generation~\cite{johnson2018image} and robotics~\cite{sanchez2018graph}. Message passing algorithms have been redefined as various graph convolution operations \cite{herzig2018mapping,kipf2016semi}. 
The graph convolutional operation is essentially equivalent to ``non-local operation"~\cite{wang2018non} derived from the self-attention concept~\cite{vaswani2017attention}. In this paper, we take ``non-local operation" as our graph convolutional operation.

\textbf{Autonomous Driving.}
Deep learning has recently been applied to learn autonomous-driving policies~\cite{chen2015deepdriving,bojarski2016end}.
Collision avoidance is an important goal for self-driving systems~\cite{kahn2017uncertainty}.
Collision vision data is difficult to collect in the real world since these are unexpected rare events.
\cite{kim2019crash} tackles the collision data scarcity by simulation. However synthetic data is still very different from real data, and hence training on simulation is not always sufficient.
In this paper, we introduce a challenging collision dataset based on real-world dashcam data. At the same time, we propose a model suitable for classifying rare events by modeling the key object interactions with limited training examples.

\begin{figure*}[t!]
    \begin{center}
        \includegraphics[width=\linewidth]{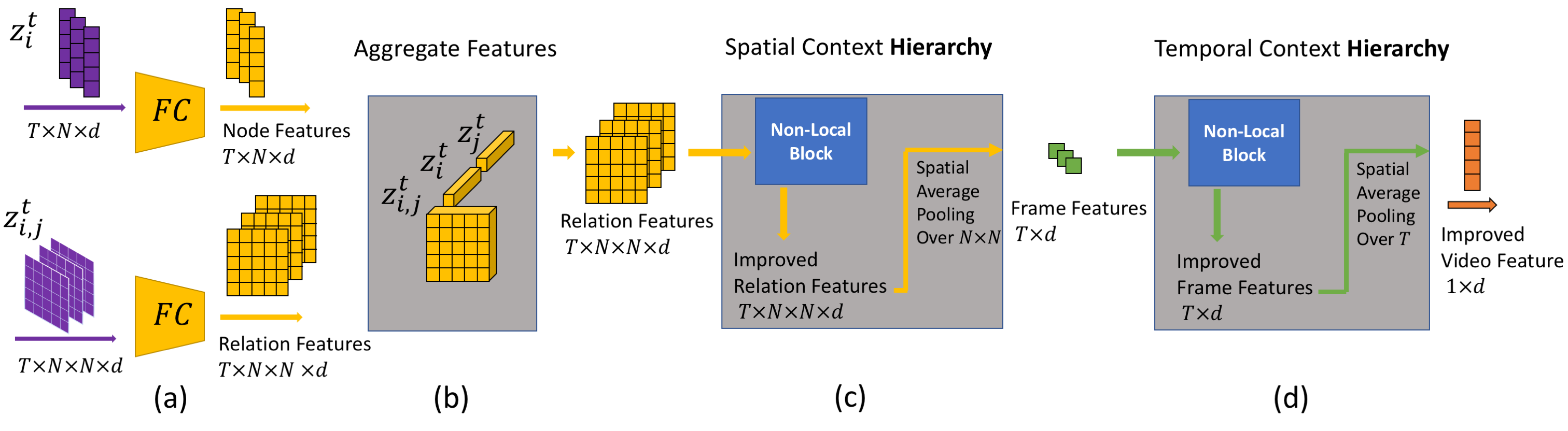}
    \vspace{-15pt}
    \caption{\small{STAG Module: 
    (a) First, the initial node features $\ve_{i}^{t}$ and union box relation features $\ve_{i,j}^{t}$ are embedded into a feature space with dimension $d$. (b) The pair of node features and their relation features are concatenated to create an aggregated representation with objects and their interactions. These are then embedded into dimension $d$ using a fully-connected network. (c) The {\bf Spatial Context Hierarchy Stage} uses a Non-Local block to obtain improved relation features and then
    pools over the spatial dimension to obtain per-frame features. (d) Then, the {\bf Temporal Context Hierarchy Stage} uses a Non-Local block to calculate improved frame features and pools over the temporal dimension. The final output is a video-level feature containing the entire scene information and video dynamics.}}
    \vspace{-15pt}
    \label{fig:actions_graph}
    \end{center}
\end{figure*}

\section{Spatio-Temporal Action Graphs}
\label{sec:model}
In this section we describe our proposed Spatio-Temporal Action Graph Network (STAG). 
\comment{
with object interaction modeling to improve video recognition. 
Our main challenge was doing a dimension reduction of the 3-D input to 1-D compact vector, in a way that mixed information from both the spatial and temporal space.
While the global pooling of convolutional filter output might ignore some crucial features for discriminating actions, object interaction modeling can compensate and capture important clue inside the video clip.
\roeih{It's unclear need to rephrase it}
But general object interactions with attention for the whole video might still miss important things happening inside each frame.
We disentangle the object spatial interaction and temporal interaction into two hierarchies, and do attention first within each frame and then temporally between frames.
By the structural attention hierarchies disentangling spatial and temporal interactions, we expect STAG can attend to important factors in each frame for better action classification.
}
The overall architecture is shown in \figref{fig:system} and the STAG module is further described in \figref{fig:actions_graph}.
%

We begin with some definitions. The following constants are used: $T$ is the number of frames, $N$ is the maximum number of objects (i.e., bounding boxes) per frame, and $d$  is the feature dimensionality (i.e., the dimension of bounding boxes descriptors). We also use $[T]$ to denote the set of input frames $\{1,\ldots,T\}$ and $[B]$ to denote the set of bounding boxes in each frame $\{1,\ldots,N\}$. At a high level the model proceeds in the following stages:
\newline
     {\bf Detection Stage} - The image is pre-processed with a detector to obtain features for each bounding box (i.e., objects) and each pair of boxes (i.e., relations).
\newline     
     {\bf Spatial Context Hierarchy Stage} - Each relation feature is refined using context from other relations, and all relation features are summarized in a single feature per frame.
\newline     
     {\bf Temporal Context Hierarchy Stage} - Each frame feature is refined using context from other frames, and all frame features are summarized in a single feature, which is then used for classification.
\subsection{Detection Stage}
\label{sec:model:detector}
Before applying the two disentangled spatial and temporal context aggregations, we construct one initial graph representation encoding objects and their relations. As a first step, we detect region proposal boxes and extract corresponding box features through an RoIAlign layer of a Faster R-CNN~\cite{faster_rcnn}. Instead of only using object features with their bounding boxes, we believe that the spatial relation of each pair of bounding box should be important and encoded into the initial graph representation for subsequent spatial and temporal context aggregation. Specifically, for each pair of boxes, we consider its union (see \figref{fig:system}) and use an RoIAlign layer to extract the union box features as initial relation features with the union boxes capturing the spatial appearance of each pair of objects.
This results in two sets of tensors shown in \figref{fig:actions_graph}a: 
\begin{itemize}
    \item Single-object features: For each time step $t\in[T]$ and box $i\in[B]$ we have a 
    feature vector $\ve_{i}^t\in\reals^d$ for the corresponding box. The feature contains the output of the RoIAlign layer.
    \item Object-pair features: For each time step $t\in[T]$ and box-pair $i\in[B],j\in[B]$ we have a feature vector $\ve_{i,j}^t\in\reals^d$ for the corresponding pair of boxes. The feature contains the output of the RoIAlign layer. 
\end{itemize}
We thus have a tensor of size $T\times N \times d$ for single object features, and a tensor of size $T\times N\times N \times d$ for object-pair features. Then (\figref{fig:actions_graph}b) we concatenate each relation (object-pair) feature with the two corresponding node features and embed the result into dimension $d$ (using an FC layer) to form one aggregated representation with objects and their interactions. The resulting tensor is thus of size $T\times N\times N \times d$.

\subsection{The STAG Module}
\label{sec:model:actions_net}
The output of the Detection Stage is a tensor of size $T\times N\times N \times d$ with features for each object interaction in each frame.  In what follows, we describe how these are refined and reduced in the two complementary hierarchies of space and time.
Before we introduce these stages, we first recap the non-local operation from~\cite{wang2018non}. These are an efficient, simple and generic component for capturing long range dependencies. Formally, given a set $\cV$ of vectors $\vv_1,\ldots,\vv_k$, the non-local operator transforms these into a new set $\cV'$ of vectors $\vv'_1,\ldots,\vv'_k$ via the function:
\begin{equation}
\label{eq:nonlocal}
\vv'_i = \frac{1}{\mathcal{C(\cV)}} \sum_{\forall j}f(\vv_{i}, \vv_{j})g(\vv_{j})
\end{equation}
where $C(\cV)$ is a normalization factor and $f$ and $g$ are learned pairwise and singleton functions.
\comment{
We define the generic operation for our edge-graph as the following:
\begin{equation}
\label{eq:edges_nonlocal}
\vect{y}_{i,j} = \frac{1}{\mathcal{C(\ve)}} \sum_{\forall k, l}f(\ve_{i, j}, \ve_{k, l})g(\ve_{k, l})
\end{equation}
while $(i,j)$ is the index of a given edge box descriptor whose response to all other $(k, l)$ edges descriptors per frame. $\ve$ is the input signal and $\vect{y}$ is the output of signal of the same size as $\ve$. A pairwise function $f$ represents the interactions between edges over time between each edge $(i,j)$ to all other edges $(k, l)$. The unary function $g$ computes a representation of the input signal of the edge feature. 
In addition, we also use the generic operation for over the temporal dimension the same as the original non-local operation as in \cite{wang2018non}:
\roeih{The second formula is not clear. You are using the same notations but it can't be the same functions.}
\begin{equation}
\label{eq:temp_nonlocal}
\vect{y}_i = \frac{1}{\mathcal{C(\ve)}} \sum_{\forall j}f(\ve_i, \ve_j)g(\ve_j)
\end{equation}
}
We next describe the final two stages of the STAG model.

\begin{figure*}[t!]
    \begin{center}
        \includegraphics[width=1.0\linewidth]{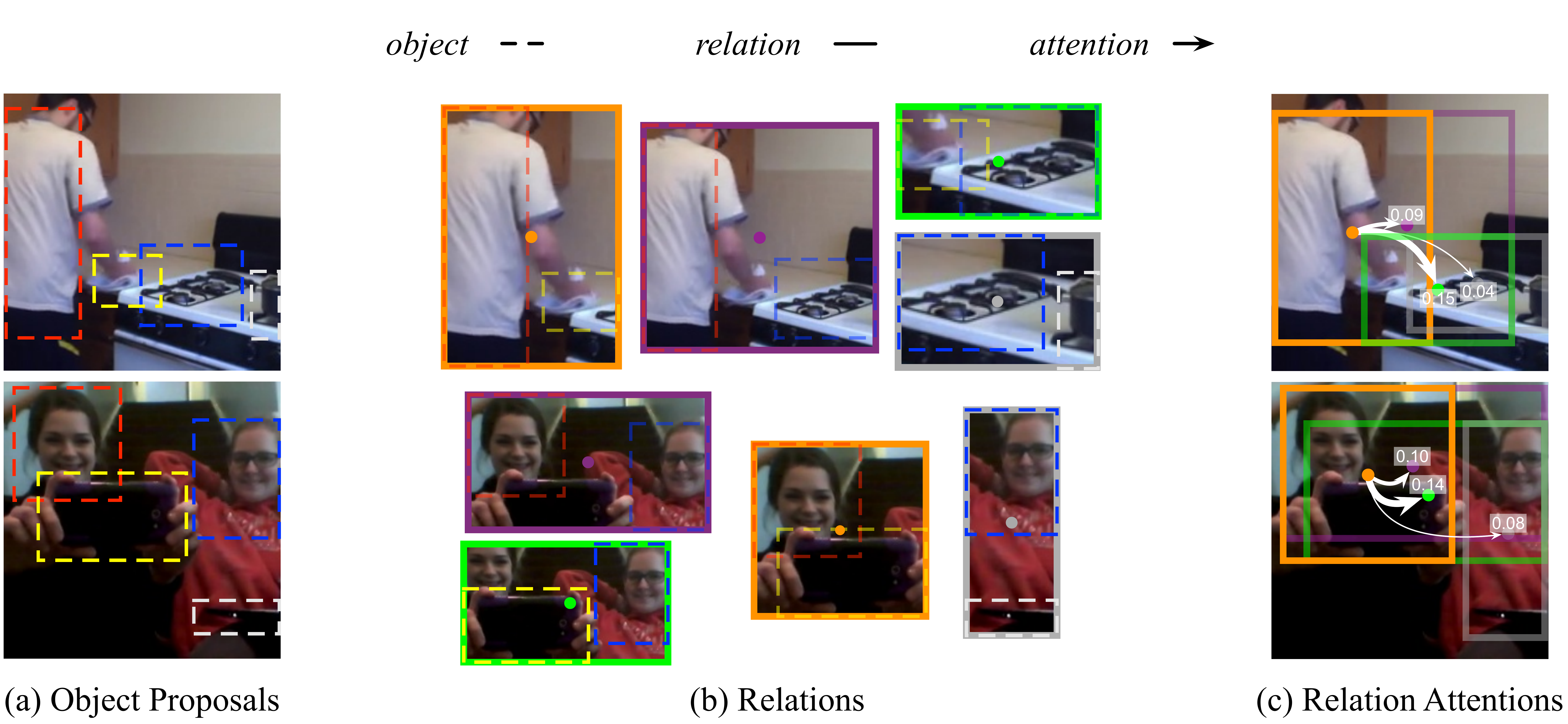}
    \vspace{-20pt}
    \caption{Spatial context hierarchy processing. Our model first extracts object proposals (see \figref{fig:actions_graph}b) and then composes them into relations by taking the union of each possible object-pair proposal pair. Our model then infers the per-frame relation interactions via an attention step. (c)  illustrates the attention on the top-3 relation-to-relation interactions and their scores for the relation box colored by orange. 
    }
    \vspace{-20pt} 
    \label{fig:charades_space}
    \end{center}
\end{figure*}

\textbf{Spatial Context Hierarchy Stage.}
The goal of this stage is twofold. First, it refines the relation features so that each feature incorporates information from 
all the other relations. This is done by applying a non-local operation to all the $N^2$ feature vectors (each of dimension $d$) that are the output of the Detection Stage. The outcome is another tensor of size $T\times N\times N \times d$. Next, it generates a single feature representing the relation information in the frame, by average pooling the above tensor, resulting in a tensor of size $T \times d$. See \figref{fig:actions_graph}c. We visualize some of the object proposals and their relations in \figref{fig:charades_space}.

\begin{figure*}[t!]
    \begin{center}
        \includegraphics[width=\linewidth]{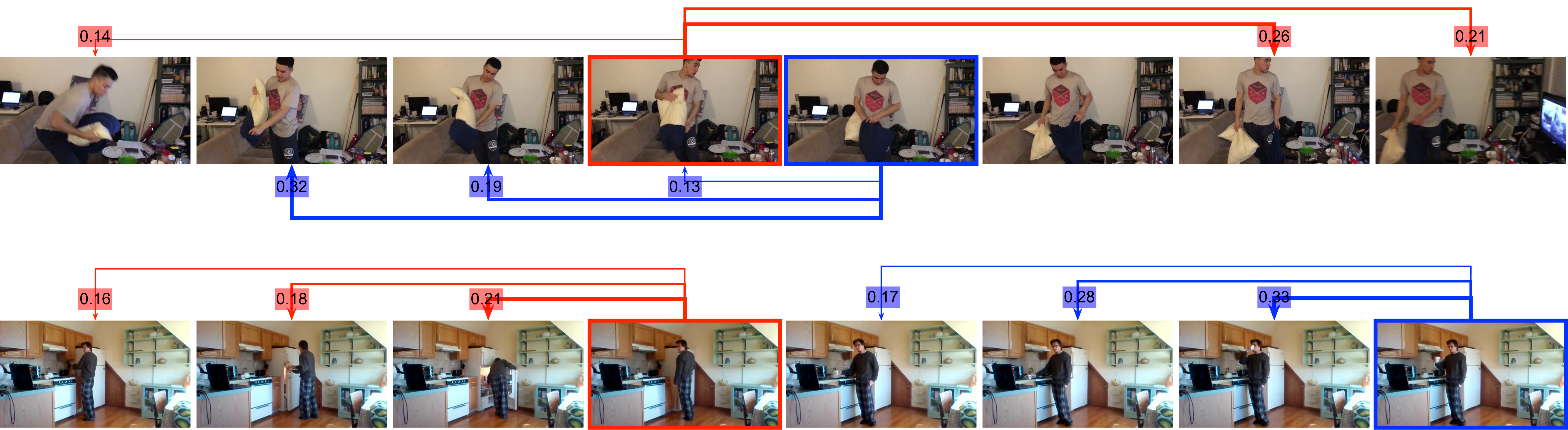}
    \vspace{-15pt}
    \caption{Our Temporal Context Hierarchy. After aggregating spatial features to represent each frame separately (see \figref{fig:actions_graph}c), our model learns relations across the frames (see \figref{fig:actions_graph}d). In each row, we show eight consecutive frames in a video in which two frames are highlighted in red or blue. We use arrows to represent the temporal attention across frames which started from either of highlighted frames. The attention score associated with them are noted beside arrows.} 
    \vspace{-20pt}
    \label{fig:charades_time}
    \end{center}
\end{figure*}

\textbf{Temporal Context Hierarchy Stage.}
At this stage, information from all frames is integrated into a single vector. This is done by applying a Non-Local block to 
the $T$ vectors (each of dimension $d$) that are the output of the Spatial Context Hierarchy Stage. The final output is a single $d$ dimensional feature vector capturing whole video information obtained by average pooling the above tensor. See \figref{fig:actions_graph}d. We visualize some of the frames and their relations in \figref{fig:charades_time}.

\section{The \text{Collision} Dataset}
\label{sec:dataset}

\begin{figure*}[t!]
\begin{center}
\includegraphics[width=\linewidth]{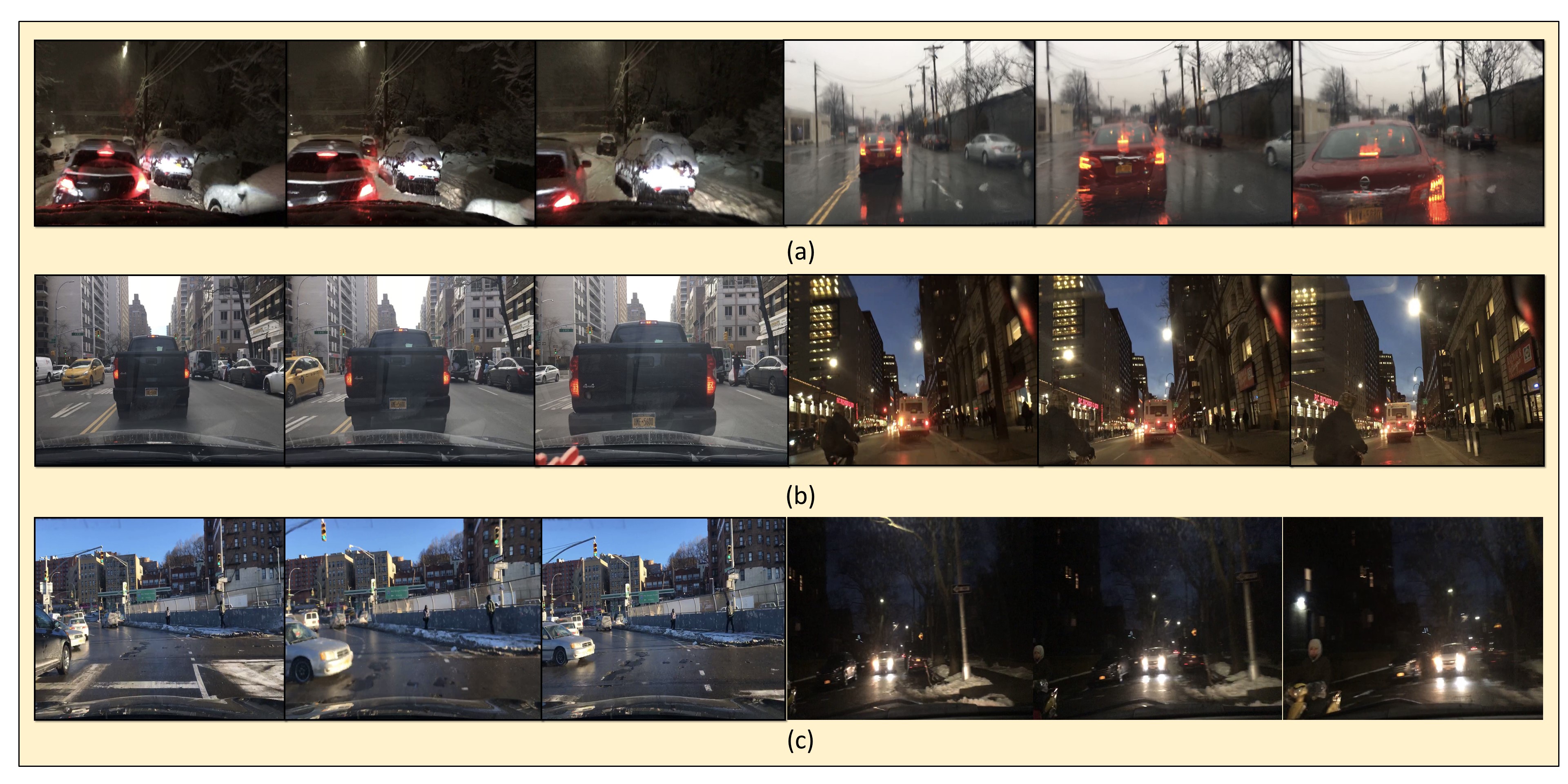}
\caption{Example visualization from the collision dataset: (a) extreme weather conditions such as snow and heavy raining. (b) near-collision with a truck/a bicycle rider. (c) day and night collisions.}
\label{fig:dataset_figures}
\vspace{-20pt}
\end{center}
\vspace{-5pt}
\end{figure*}

We introduce the Collision dataset comprised of real-world driving videos. Such a dataset is valuable for developing autonomous driving models. Using videos, and specifically visual information, is important for accurate and timely prediction of collisions. The dataset contains rare collision events from diverse driving scenes, including urban and highway areas in several large cities in the US. These events encompass collision scenarios (i.e., scenarios involving the contact of the dashcam vehicle with a fixed or moving object) and the near-collision scenarios (i.e., scenarios requiring an evasive maneuver to avoid a crash). Such driving scenarios most often contain interactions between two vehicles, or between a vehicle and a bike or pedestrian.
Classifying such events therefore naturally requires modeling object interactions, which was our motivation for developing the STAG model.
We will release a publicly available challenge based on this dataset upon acceptance of paper.

\begin{table}[t!]
\footnotesize
\centering
\begin{tabular}[t]{|l |c |} 
\hline
Party type & dist. \\  \hline
Vehicle & 85\% \\ \hline
Bike &  6\%\\ \hline
Pedestrian &  6\% \\ \hline
Road object &  1\% \\ \hline
Motorcycle &  1\% \\ \hline
\end{tabular}
\hfill
\begin{tabular}[t]{|l| c |} 
\hline
Weather & dist. \\ \hline
Clear & 93\% \\ \hline
Rain &  5.3\% \\ \hline
Snow &  1.7\% \\ \hline
\end{tabular}
\hfill
\begin{tabular}[t]{|l| c |} 
\hline
Lighting & dist. \\ \hline
Day & 62\% \\ \hline
Night &  38\% \\ \hline
\end{tabular}
\hspace{2.0cm}
\caption{Collision dataset statistics: involved party, weather, and lighting conditions.}
\label{tbl:dataset-stats}
\vspace{-15pt}
\end{table}

{\bf Data collection.}  The data was collected from a large-scale deployment of connected dashcams. Each vehicle is equipped with a dashacam and a companion smartphone app that continuously captures and uploads sensor data such as IMU and gyroscope readings. Overall, the vehicles collected more than 10 million rides, and rare collision events were automatically detected using a triggering algorithm based on the IMU and gyroscope sensor data.\footnote{The algorithm is tuned to capture driving maneuvers such harsh braking, acceleration, and sharp cornering.} The events were then manually validated by human annotators based on visual inspection to identify edge case events of collisions and near-collisions, as well as non-risky driving events.  Of all the detected triggers, our subset contains 743 collisions and 60 near-collisions from different drivers. 
Each video clip contains one such event typically occurring in the middle of the video clip.
The clip duration is approximately 40 seconds on average and the frame resolution is $1280 \times 720$.

{\bf The full and few-shot datasets.} 
We created two data versions. 
The full dataset contains a total of 803 videos with 732 videos as training data (44 of them are near collision) and 71 videos as test data (6 of them are near-collision). 
We use a relatively low frame rate of 5 fps to convert video clips to frames, in order to avoid using near-duplicate frames.
Each clip is broken into three segments to train our model.
Specifically, we split each video into three segments of 20 frames each: two negative segments (non-risky driving scenes) and one positive segment (a collision scene) for each collision event. The positive segment is sampled at the time of collision, and the two non-overlapping negative segments are sampled before the time of collision, since after collision the scene is already in a collision state. For the near-collision event, we sample three negatives since there is no positive segment in the video clip. After this processing, we have a total 2409 video segments, out of which 1656 are negative examples and 753 are positives. The few-shot dataset is purposely designed to motivate the development of few-shot recognition algorithms. It contains 125 videos, with 25 training videos and 100 testing videos. Data processing is the same as the full-version. The positive to negative ratio in both versions of dataset is approximately $1:3$.

{\bf Diversity.} Our dataset is collected for recognizing collisions in natural driving scenes. To get an intuitive feeling for our collision dataset, we visualize several video examples in ~\figref{fig:dataset_figures}. The coverage of the dataset includes various types of weather conditions (\figref{fig:dataset_figures}a), the parties involved (\figref{fig:dataset_figures}b) and lighting conditions (\figref{fig:dataset_figures}c). 
We use visual inspection to analyze the identity of parties involved in the collisions in \tabref{tbl:dataset-stats}. We find that most of the data (85\%) consists of crashes involving two vehicles, and collisions involving pedestrians and cyclists takes up to 6\% each. \tabref{tbl:dataset-stats} also shows the distribution of weather conditions and lighting conditions. With a majority of clear weather (93\%), the extreme rain and snow video clips take 5.3\% and 1.7\% each. Day-time takes 62\% with the rest of 38\% being night-time.



\section{Experiments on the Collision Dataset}
\label{sec:exper}

Our method is designed to address rich inter-object interactions. The only dataset that captures these as Charades, whereas other datasets contain limited human-object interactions. It was thus natural to evaluate it on Charades and Collision. We next describe the application of our STAG model to those datasets.

\subsection{Implementation Details}
\label{sec:exper:details}

\noindent \textbf{Model Details.} 
We use Faster R-CNN with ResNet50 as a backbone, taking a sequence of $T=20$ frames, and generating bounding box proposals for each of the $T$ frames. Specifically, the strides in FPN are set as the same as \cite{fpn}.

The input frames are resized to the maximum dimension of 256 with padding. Considering the training time and memory limit, we take the top $N=12$ region proposals on each frame after non-maximum suppression with IoU threshold 0.7, which are sufficient for capturing the semantic information on the Collision dataset. Features for the $N=12$ objects and $N\cdot N$ object interaction relations are extracted following~\secref{sec:model}, resulting in feature representations $\zz_i$ for objects and $\zz_{i,j}$ for relations.

\noindent \textbf{Training and Inference.} 
We train STAG using SGD with momentum $0.9$ and an initial learning rate $0.01$. The learning rate is decayed by a factor of $0.5$ each epoch, and gradient is clipped at norm $5$. Each batch includes a video segment of $T$ frames.
Two kinds of ground truth data are utilized during training: the ground truth bounding box annotations on each frame and the collision label per segment.

The loss for the STAG model contains two components: the bounding box localization related losses used in the Faster-RCNN detector and the multi-class activity classification loss, as is standard with two stage detectors. To train our STAG model for collision prediction, we apply a binary cross entropy loss between the binary collision prediction logit and the ground-truth collision label.

\subsection{Model Variants}
\label{sec:exper:ablations}
The STAG model progressively processes the box features and the spatial appearance features of pairwise boxes $\zz_i^t,\zz_{i,j}^t$ into a single vector for final activity recognition. To explore the importance of the spatial and temporal aspects of STAG, we consider the following variants:
\newline
(1) \texttt{LSTM Spatial Graph} - We study the effect of the STAG ``Temporal Context Hierarchy'' stage, as  compared to a recurrent neural network based solution. To do so, we replace the ``Temporal Context Hierarchy'' stage with an LSTM that processes the same tensor of size $T\times d$.
\newline
(2) \texttt{LSTM Boxes} - We study the effect of the ``Spatial Context Hierarchy'' stage by replacing it with average pooling of the node features, to obtain a tensor of size $T\times d$. We also train two other popular activity recognition models on Collision dataset: the C3D model~\cite{tran2015learning} and I3D model~\cite{Carreira2017QuoVA}.
We used pretrained weights for C3D and I3D. The C3D was pretrained on Sports-1M while the I3D was pretrained on Kinetics.


\begin{table}[t!]
\begin{center}

\label{results}
\small
   \begin{tabular}{|l|cc|} 
     \hline
     \multicolumn{1}{|c|}{} &    \multicolumn{2}{c|}{Accuracy} \\
     {} & {Full Dataset} & {Few-shot Dataset} \\
     \hline
     I3D & 82.4 & 76 \\ 
     C3D & 79.9 & 72 \\ 
     LSTM Spatial Graph & 77.5 & 67 \\
     LSTM boxes & 69.5 & 69 \\
     \hline
     \textsc{STAG} & \textbf{84.5} & \textbf{76.3} \\
     \hline
\end{tabular}
  \vspace{-10pt}
  \caption{Classification accuracy on the Collisions dataset for the STAG model and its variants, and the C3D \& I3D model.}
  \vspace{-30pt}
  \label{sec:exper:results}
\hspace{2.0cm}
\end{center}
\end{table}


\begin{table}[t!]
\begin{center}
\label{results}
\small
  \begin{tabular}{|l|cc|}
     \hline
     \multicolumn{1}{|c|}{} &
     \multicolumn{2}{c|}{Accuracy} \\
     {} & {Full Dataset} & {Few-shot Dataset} \\
     \hline
     LSTM spatial Graph & 83.56 & 73.1 \\
     LSTM boxes & 81.2 & 72.3 \\
     \hline
     \textsc{STAG} & \textbf{85.5} & \textbf{76.7} \\
     \hline
\end{tabular}
  \vspace{-10pt}
  \caption{Classification accuracy on the Collisions dataset for STAG model and variants, when averaged with the C3D model.}
  \vspace{-20pt}
  \label{sec:exper:fusion_results}
\hspace{2.0cm}
\end{center}
\end{table}

\subsection{Results}
We first compare the STAG results on the full dataset to the model variants described in \secref{sec:exper:ablations}. ~\tabref{sec:exper:results} reports classification accuracy. Firstly, STAG outperforms all the other models including C3D and I3D. Replacing the temporal processing in STAG with an LSTM as in \textsc{LSTM Spatial Graph}, we get 7\% accuracy decrease, showing the superiority of our temporal modeling over LSTM. Further removing the pairwise object modeling, we see accuracy further decrease by 8\% in \textsc{LSTM Boxes}. 

Finally, we consider a simple ensemble model of STAG and C3D by simply averaging their output scores. Results of this combination are shown in \tabref{sec:exper:fusion_results}. We can see the combination improves the original C3D accuracy, showing the benefits of object interaction modeling. Among all the ensemble results, the STAG model still maintains the highest accuracy result 85.5\%.

We also show the results on the few-shot dataset in \tabref{sec:exper:results} and \tabref{sec:exper:fusion_results}. It can be seen that the two LSTM model variants almost fail on this challenging dataset. Although our STAG model achieves marginal improvement compared to the C3D and I3D, the relative low accuracy numbers highlight the challenges of this setting.\footnote{We note however, that all these models are not specifically designed for the few-shot setting.} We encourage the community to further develop few-shot based activity recognition models to tackle this challenging few-shot dataset.

\comment{
To see the complementary strengths of STAG with hierarchical object interaction modeling to C3D, we fuse the prediction from C3D with STAG model and its variants \textsc{LSTM Spatial Graph} and \textsc{LSTM Boxes}.}


\subsection{Ablation Studies}
\label{sec:exp:collision_ablation}
We also design some direct ablation studies for the components in our STAG model. To validate the effectiveness of our disentangled spatio-temporal hierarchies, we design two ablation studies for the two attention hierarchies: 
(1) \texttt{STAG Space} - Replacing the spatial hierarchy by directly pooling.
(2) \texttt{STAG Time} - Replacing the temporal hierarchy by directly pooling.

The results are shown in \tabref{sec:exper:hier_ablations}. It can be seen that both ablations decrease accuracy, but that the temporal hierarchy has a larger effect on performance.

In addition to our visual appearance relation features, we explore the use of different relation features in ~\tabref{sec:exper:hier_ablations}: 
(1) \texttt{STAG Cat} - Set edge feature to be just the concatenation of the corresponding node features (i.e., union box is not used).
(2) \texttt{STAG Sim} - Set edge feature to be cosine similarity of the two corresponding node features (see \cite{Wang_videogcnECCV2018}).

Both methods result in approximately one point accuracy decrease, indicating the superiority of using spatial appearance features of union boxes as edge features in our hierarchical STAG models.

\begin{table}[t] 
    \small
    \centering
    {
    \begin{tabular}{lllc}
        & \textbf{Edge} & \textbf{Hierarchy} & \textbf{Accuracy} \\
        \toprule
        STAG Cat & Node concat. & Space \& Time & 83.1 \\
        STAG Sim & Cosine sim. & Space \& Time & 83.5 \\
        \midrule
        STAG Time & Node interactions & Time only & 78.8 \\
        STAG Space & Node interactions & Space only & 82.6 \\
        \textbf{STAG} & Node interactions & Space \& Time & \textbf{84.5} \\
        \bottomrule
        \addlinespace
    \end{tabular}
    }
    \vspace{-10pt}
    \caption{Hierarchy \& Edge features Ablations on the Collision dataset. ``Node Interactions'' refers to using relations features for the edge features.}
    \vspace{-10pt}
    \label{sec:exper:hier_ablations}
\end{table}

\section{Experiments on the Charades Dataset}
\vspace{-5pt}

To further validate the effectiveness of our model on publicly available action recognition benchmarks, we also evaluate it on the Charades dataset~\cite{sigurdsson2016hollywood}. We follow the official split (8K training and 1.8K validation videos) to train and test our model.
The average video duration is around 30 seconds with 157 multiple action classes and we report our results by the metric of mean Average Precision (mAP).

We follow the same experiment setup as described in STRG (Spatio-Temporal Region Graph)~\cite{Wang_videogcnECCV2018} and use a backbone network of ResNet-50 Inflated 3D ConvNet (I3D) \cite{wang2018non} for all of our experiments. 

\noindent \textbf{Training and Inference.} 
Our network takes 32 video frames as inputs which are sampled at 6fps, resulting in maximum input duration of about 5 seconds. We use a spatial resolution of 224~$\times$~224. Data augmentation is as in \cite{simonyan2014very}. The top $N=15$ object proposals are selected.

To train our model, we follow the same training schedule as specified in STRG using a mini-batch of 8 videos for each iteration and repeat it with 100K iterations in total. The training objective is a simple cross entropy loss. During inference, we apply multi-crop testing~\cite{wang2018non,Wang_videogcnECCV2018} for better performance and the final recognition results are based on late fusion of classification scores.

\noindent \textbf{Results.} 
\tabref{exp:charades} compares STAG to various baselines on Charades. It can be seen that it compares favorably with prior works that used the same ResNet-50 I3D backbone: STAG improves 5.4\% over the I3D model, and 1.0\% over STRG. 

\noindent \textbf{Ablations.} 
Next, we run the same ablation studies (\texttt{STAG Space},  \texttt{STAG Time},  \texttt{STAG Cat}) as in \secref{sec:exp:collision_ablation}. Results are shown in \tabref{tab:charades_abl}. It can be seen that as with the Collisions data, all STAG ablations result in decreased accuracy. \texttt{STRG Sim} refers to the STRG model which uses cosine similarity between the nodes as the edge features, while either not discriminating the nodes from different frames at all or only applying heuristic backward-forward node association as space-time hierarchy. We compare \texttt{STRG Sim} to \texttt{STAG Relation}, a model that uses the same relation features as STRG, and thus it is a direct comparison between the similarity from \cite{Wang_videogcnECCV2018} and  our relation features approach. Our design of relations feature as the edge feature  captures object interactions and brings a 0.6\% performance gain over \texttt{STRG Sim}.

\begin{table}[t] 
    \centering
    \resizebox{\linewidth}{!}{%
    \begin{tabular}{lllc}
        & \textbf{Backbone} & \textbf{Modality} & \textbf{mAP}
        \\
        \toprule
        2-Steam~\cite{simonyan2014two} & VGG-16 & RGB w/ Flow & 18.6
        \\
        2-Steam w/ LSTM~\cite{simonyan2014two} & VGG-16 & RGB w/ Flow & 18.6
        \\
        Async-TF~\cite{sigurdsson2017asynchronous} & VGG-16 & RGB w/ Flow & 22.4
        \\
        a Multiscale TRN~\cite{zhou2018temporal} & Inception & RGB & 32.9
        \\
        I3D \cite{Carreira2017QuoVA} & Inception & RGB & 32.9
        \\
        \midrule
        I3D \cite{Wang_videogcnECCV2018} & R50-I3D & RGB & 31.8
        \\
        STRG \cite{Wang_videogcnECCV2018} & R50-I3D & RGB & 36.2
        \\
        \textbf{STAG (ours)} \cite{wang2018non} & R50-I3D & RGB & \textbf{37.2}
        \\
        \bottomrule
        \addlinespace
    \end{tabular}
    }
    \vspace{-14pt}
    \caption{Classification mAP in the Charades dataset. \cite{sigurdsson2016hollywood}}
    \vspace{-10pt}
    \label{exp:charades}
\end{table}


\begin{table}[t] 
    \centering
    \resizebox{\linewidth}{!}
    {
    \begin{tabular}{lllc}
        & \textbf{Edge} & \textbf{Hierarchy} & \textbf{mAP} \\
        \toprule
        I3D & - & - & 31.8 \\
        \midrule
        STRG Sim~\cite{Wang_videogcnECCV2018} & Cosine sim. & No hierarchy & 35.0 \\
        STRG \cite{Wang_videogcnECCV2018} & Cosine sim. & Space-Time Heuristic & 36.2 \\
        \midrule
        STAG Relation & Node interactions & No hierarchy & 35.6 \\
        STAG Cat & Node concat. & No hierarchy & 34.5 \\
        \midrule
        STAG Space & Node interactions & Space only & 34.7 \\
        STAG Time & Node interactions & Time only & 36.6 \\
        \textbf{STAG} & Node interactions & Space \& Time & \textbf{37.2} \\
        \bottomrule
        \addlinespace
    \end{tabular}
    }
    \vspace{-14pt}
    \caption{Hierarchy \& Edge features Ablations on the Charades dataset.}
    \vspace{-20pt}
    \label{tab:charades_abl}
\end{table}
\section{Conclusion}
\label{sec:conclusion}

The interaction of objects over time is often a critical cue for understanding activity in videos.
We presented a novel inter-object graph representation which included explicit appearance models for edge-terms in the graph as well as a novel factored embedding of the graph structure into spatial and temporal representation hierarchies.
We demonstrated the effectiveness of our model on the Charades activity recognition dataset as well as on a new dataset of driving near-collision events; our model significantly improved performance compared to baseline approaches without object-graph representations or with previous graph-based models.

\vspace{-5pt}
\section*{Acknowledgements}
\label{sec:ack}
\vspace{-5pt}
This work was completed in partial fulfillment for the Ph.D degree of the first author.

{\small
\bibliographystyle{ieee}
\bibliography{egbib}
}

\appendix
\newpage
\section*{Supplementary Material}
\label{sec:supp}

This supplementary material includes: (1) Model details for the Charades data experiments, (2) Model details for the Collision data experiments, (3) Additional qualitative results for the Collision data.

\section{Model Details for the Charades Experiment}
\noindent \textbf{Backbone Architecture.} \quad 
We follow \cite{Wang_videogcnECCV2018} and use the ResNet-50 I3D model as the backbone for all of our models. Our backbone configuration is described in \tabref{tab:r50}. For the ResNet-50 I3D baseline model, we reshape the final pooled feature map to be a 2048-dimensional feature vector and apply a simple fully-connected layer for classification.

\noindent \textbf{Region Proposal Network.} \quad 
For object proposals, we use the Region Proposal Network (RPN) from~\cite{faster_rcnn, Detectron2018} which was pretrained on the MSCOCO object detection dataset~\cite{lin2014microsoft}. Specifically, we use the RPN with ResNet-50 and an FPN~\cite{lin2017feature} backbone. It should be noted that the proposals used in our model are class-agnostic.

First, we sample each selected frame from two consecutive frames and use the RPN to extract proposal boxes on the dense feature maps from the extracted original video clips after the res$_5$ layer. Next, we project these onto our feature map coordinates for later RoIAlign operations~\cite{he2017maskrcnn}. Finally, each box region is mapped to a $7 \times 7 \times 2048$ feature map which is then max pooled to a feature vector representing that region.


\section{Model Details for the Collision Experiment}

\begin{table}[t!] 
    \centering
    \resizebox{\linewidth}{!}
    {
    \begin{tabular}{lcc}
        \textbf{Layer} & \textbf{Configuration} & \textbf{Output size} \\
        \toprule
        input & - & $32 \times 224 \times 224$ \\
        \midrule
        conv$_1$ &
        $5 \times 7 \times 7, 64, \text{stride }1, 2, 2$ &
        $32 \times 112 \times 112$
        \\
        pool$_1$ &
        $1 \times 3 \times 3, \text{max}, \text{stride }1, 2, 2$ &
        $32 \times 56 \times 56$
        \\
        \midrule
        res$_2$ &
        $\begin{bmatrix}
        3 \times 1 \times 1, 64 \\
        1 \times 3 \times 3, 64 \\
        1 \times 3 \times 3, 256 
        \end{bmatrix}$ $\times 3$ & $32 \times 56 \times 56$ \\
        pool$_2$ & $1 \times 1 \times 1, \text{max}, \text{stride }2, 1, 1$ & $16 \times 56 \times 56$ \\
        \midrule
        res$_3$ &
        $\begin{bmatrix}
        3 \times 1 \times 1, 128 \\
        1 \times 3 \times 3, 128 \\
        1 \times 3 \times 3, 512 
        \end{bmatrix}$ $\times 4$ & $16 \times 28 \times 28$ \\
        \midrule
        res$_4$ &
        $\begin{bmatrix}
        3 \times 1 \times 1, 256 \\
        1 \times 3 \times 3, 256 \\
        1 \times 3 \times 3, 1024 
        \end{bmatrix}$ $\times 6$ & $16 \times 14 \times 14$ \\
        \midrule
        res$_5$ &
        $\begin{bmatrix}
        3 \times 1 \times 1, 512 \\
        1 \times 3 \times 3, 512 \\
        1 \times 3 \times 3, 2048 
        \end{bmatrix}$ $\times 6$ & $16 \times 14 \times 14$ \\
        pool$_5$ &
        $16 \times 14 \times 14, \text{avg}, \text{stride }1, 1, 1$ &
        $1 \times 1 \times 1$
        \\
        \bottomrule
        \addlinespace
    \end{tabular}
    }
    \vspace{-10pt}
    \caption{The backbone ResNet-50 I3D model~\cite{wang2018non,Wang_videogcnECCV2018} used in our paper. The layer configurations are in $T \times H \times W$ format to represent the dimensions of filter kernels.}
    \vspace{-10pt}
    \label{tab:r50}
\end{table}

\begin{figure*}[t!]
\begin{center}
\includegraphics[width=1\linewidth]{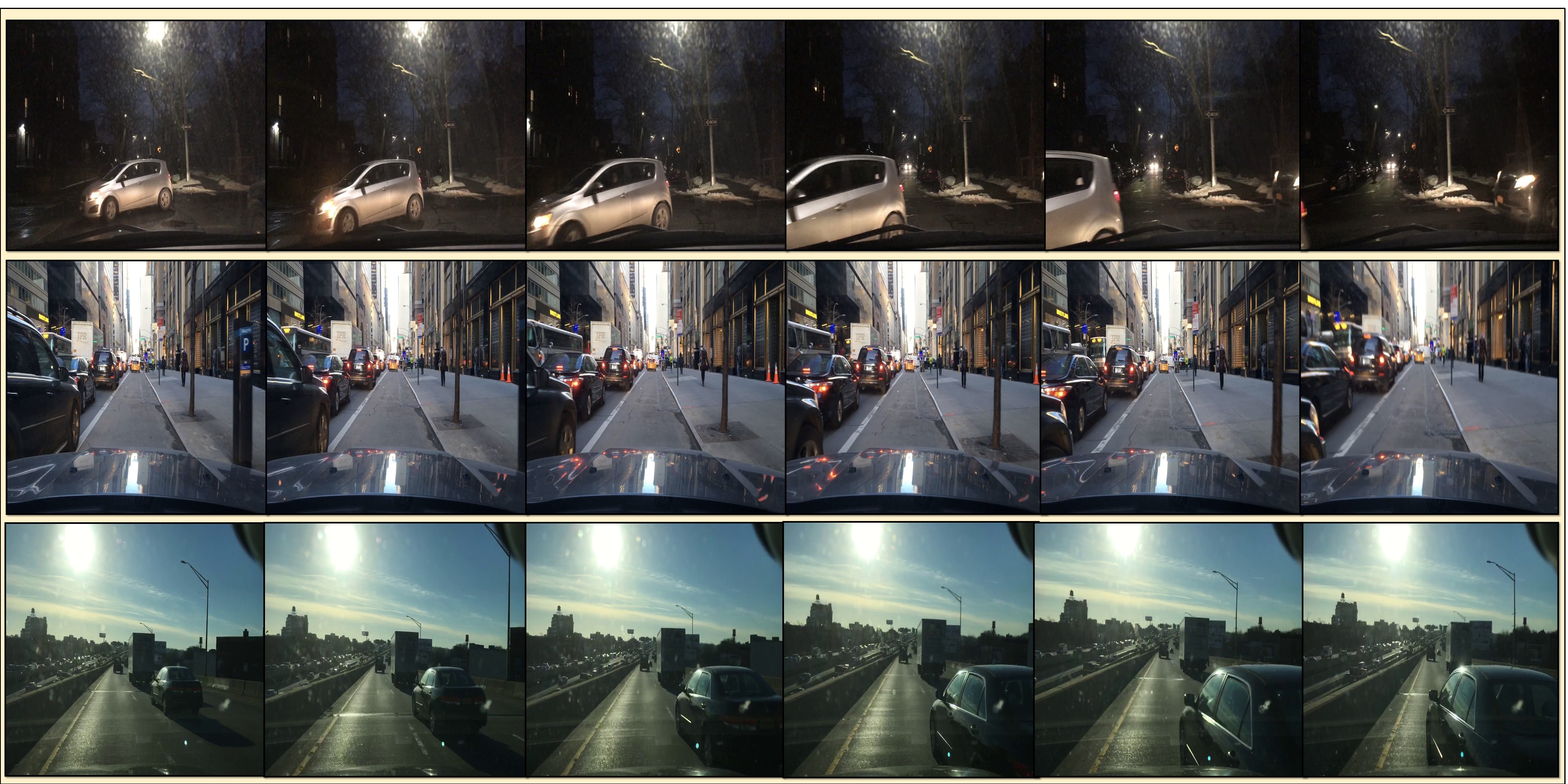}
\caption{Success and failure cases for STAG in the Collision dataset. {\bf Top Row}: a near-collision example (i.e., the correct label is ``no collision'') 
where the STAG model correctly classifies, and the C3D and I3D models do not. {\bf Middle Row}: a collision example (i.e., the correct label is ``collision''), where the C3D and I3D models are correct, while the STAG model errs. {\bf Bottom Row}: a collision example (i.e., the correct label is ``collision''), where the STAG model trained on the Few-shot dataset makes a correct prediction.}
\label{fig:exper:full}
\end{center}
\end{figure*}

\begin{figure*}[t!]
    \begin{center}
        \includegraphics[width=\linewidth]{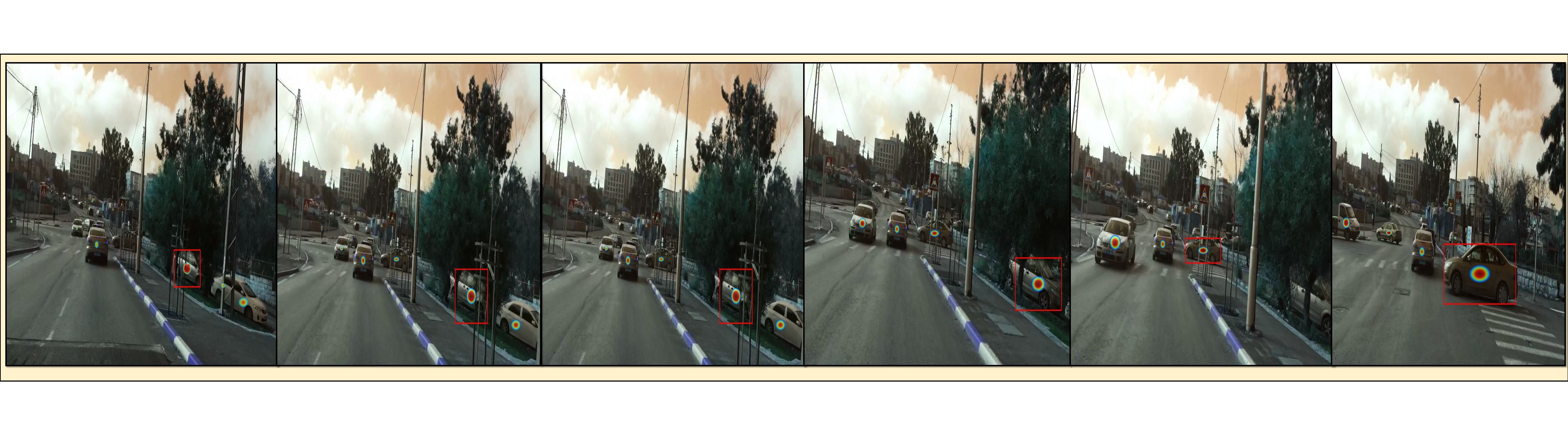}
    \caption{{\bf Explainability via Attention}. A sequence of frames, with superimposed heat-map as calculated using the spatial attention weight in each frame. The object with the highest attention score is shown in the red bounding box, and the heat-maps of the other objects describe their weights with respect to the red bounding box.} 
    \label{fig:attention_graph}
    \end{center}
\end{figure*}

\noindent \textbf{Backbone Architecture.} \quad 
We use Faster R-CNN \cite{faster_rcnn} with ResNet50 as a backbone, taking a sequence of $T=20$ frames, and generating bounding box proposals for each of the $T$ frames. Specifically, the strides in FPN are set as {(4, 8, 16, 32, 64)}, and for the RPN we set the anchor scales as {(32, 64, 128, 256, 512)}, aspect ratios as {(0.5, 1, 2)} and anchor stride as $1$. 

\noindent \textbf{Region Proposal Network.} \quad 
The RPN proposals were filtered by non-maximum suppression with IoU threshold of 0.7. 
The model then subsampled the most likely 12 ROIs from an initial 2000 ROIs of the RPN. 
Using a higher number or ROIs could potentially be a major issue for time and space complexity. However, scenes can typically be represented by only 12 objects because they capture the key objects in the scene. Features for the $12$ objects and $12\cdot 12$ relations are extracted with RoIAlign operations~\cite{he2017maskrcnn} resulting in features $\zz_i$ for objects and $\zz_{i,j}$ for relations. 
Both the objects and relations ROIs are pooled to $7 \times 7$ followed by $1 \times 1$ convolution embedding layer in the dimensions of $T \times N \times d $ and $T \times N \times N \times d$, respectively. The Faster R-CNN was pretrained on the BDD dataset \cite{Madhavan_bdd} using the default split as suggested therein. 

\section{Additional Qualitative Results for the Collision Experiment}

\noindent \textbf{Qualitative Analysis.} \quad 
Visual inspection of success and failure cases reveals an interesting pattern. We observe that STAG outperforms C3D and I3D on ``near-collision'' cases such as that in the top row of \figref{fig:exper:full}. Correctly classifying such cases requires understanding the relative configuration of the objects in order to determine if there was an incident or not. On the other hand C3D and I3D may put too much weight on events such as speed-change which are not always predictive of an accident.
The C3D and I3D models outperform STAG on cases where objects are not clearly visible. For example in the middle row of \figref{fig:exper:full}.
The bottom row of \figref{fig:exper:full} shows one correct collision prediction examples from our STAG model trained on the Few-shot dataset.

\noindent \textbf{Attention Visualization.} \quad 
Our STAG model uses attention across spatial and temporal hierarchies to encode a video into a single vector. One advantage of attention models is their interpretability. Specifically, one can view the attention map that the model produces to understand which parts of the input have more influence on the decision. For the case of collision detection, localizing the collision evidence is clearly an important feature.
\figref{fig:attention_graph} provides a nice illustration of the insight that attention maps provide. It can be seen that objects that pose more danger to the driver tend to receive higher attention values. The heat-maps per frame are generated using Gaussians filter with attention scores as confidence, learned by the spatial hierarchy. The object with the highest attention score is marked as the red bounding box, and the heat-maps of the other objects describe their weights with respect to the red bounding box.


\end{document}